\documentclass[a4paper,fleqn]{cas-dc}

\usepackage[numbers]{natbib}
\usepackage[ruled,vlined]{algorithm2e}
\usepackage{rotating}
\usepackage{subfigure}

\def\tsc#1{\csdef{#1}{\textsc{\lowercase{#1}}\xspace}}
\tsc{WGM}
\tsc{QE}

\usepackage{booktabs}
\usepackage{multirow}
\usepackage[ruled,vlined]{algorithm2e}
\usepackage{subfigure}
\usepackage{arydshln}

\usepackage{xcolor}

\makeatletter
\DeclareRobustCommand\onedot{\futurelet\@let@token\@onedot}
\def\@onedot{\ifx\@let@token.\else.\null\fi\xspace}

\makeatother
\newcommand{\name}[0]{bandLKH\xspace}

\begin{document}
\let\WriteBookmarks\relax
\def\floatpagepagefraction{1}
\def\textpagefraction{.001}

\shorttitle{Bandit based Dynamic Candidate Edge Selection in Solving TSPs}

\shortauthors{Wang et al.}  


\title [mode = title]{Bandit based Dynamic Candidate Edge Selection in Solving Traveling Salesman Problems}  

\author[1]{Long Wang}
\ead{m202273734@hust.edu.cn}
\credit{Conceptualization of this study, Methodology, Software, Writing and revision}
\cormark[1]
\cortext[cor1]{The first three authors contributed equally.}
\address[1]{School of Computer Science and Technology, Huazhong University of Science and Technology, China 430074}
\address[2]{MIS, University of Picardie Jules Verne, France 80039}

\author[1]{Jiongzhi Zheng}
\ead{jzzheng@hust.edu.cn}
\cormark[1]
\credit{Conceptualization of this study, Methodology,  Writing and revision}

\author[1]{Zhengda Xiong}
\cormark[1]
\ead{xiongzd@hust.edu.cn}
\credit{Writing and revision}

\author[2]{ChuMin Li}
\ead{chumin.li@u-picardie.fr}
\credit{Writing and revision}

\author[1]{Kun He}[orcid=0000-0001-7627-4604]
\cormark[2]
\ead{brooklet60@hust.edu.cn}
\cortext[cor2]{Corresponding author.}
\credit{Conceptualization of this study, Methodology, Supervision, Writing and revision}

\begin{abstract}
Algorithms designed for routing problems typically rely on high-quality candidate edges to guide their search, aiming to reduce the search space and enhance the search efficiency. 
However, many existing algorithms, like the classical Lin-Kernighan-Helsgaun (LKH) algorithm for the Traveling Salesman Problem (TSP), often use predetermined candidate edges that remain static throughout local searches. This rigidity could cause the algorithm to get trapped in local optima, limiting its potential to find better solutions. To address this issue, we propose expanding the candidate sets to include other promising edges, providing them an opportunity for selection. Specifically, we incorporate multi-armed bandit models to dynamically select the most suitable candidate edges in each iteration, enabling LKH to make smarter choices and lead to improved solutions. Extensive experiments on multiple TSP benchmarks show the excellent performance of our method. Moreover, we employ this bandit-based method to LKH-3, an extension of LKH tailored for solving various TSP variant problems, and our method also significantly enhances LKH-3's performance across typical TSP variants. 
~~~ ~~~
\end{abstract}




\maketitle

\sloppy{}

\section{Introduction}
\label{Sec-Intro}

The Traveling Salesman Problem (TSP) is a classic NP-hard combinatorial optimization problem. Given an undirected complete graph where the distance between each pair of vertices (i.e., cities) is known, TSP aims to find the shortest path that starts from a starting city, passes each city exactly once, and then returns to the starting city. As the basic problem of many routing problems~\cite{VRP,routing1,routing2,tsp1,cor/tsp1,cor/tsp2}, TSP also has many practical applications~\cite{cor/routing1,cor/routing2}.



Heuristic algorithms are known to be most efficient for solving the TSP, and they can be divided into two categories: global search and local search. Global search methods~\cite{eax} attempt to explore a wide solution space with a global perspective incorporating entropy in individual selection and partial crossover which are impressive in genetic algorithm~\cite{genetic1,genetic2}. However, it is hard and time-consuming for them to tackle super-large instances with huge solution spaces. 
Local search methods always maintain the current solution and explore its neighborhood space, which are efficient and suitable for instances with various scales. In this paper, we mainly focus on local search methods, among which the Lin-Kernighan-Helsgaun (LKH)~\cite{lkh} algorithm is one of the most representative and best-performing methods.



The LKH algorithm is an outstanding method among the Lin-Kernighan (LK)~\cite{lk} series that collects some promising edges in the candidate sets of each city during initialization. When using the local search operators, such as $k$-opt, to adjust the current solution, the LK-based algorithms restrict the new edges to be added to belong to the candidate sets. Therefore, the algorithm performance heavily depends on the quality of the candidate edges. LKH proposes an effective metric called $\alpha$-value calculated based on a 1-tree structure to select the candidate edges~\cite{Held1970}, resulting in its excellent performance.



However, the candidate edges in LKH are generally predetermined in the initialization stage 
and then keep unchanged during the searching process. Therefore, the content of each city's candidate edges is relatively fixed, and LKH only associates each city with about a fixed number of candidate edges (5 by default). Although the $\alpha$-value is promising in evaluating the quality of the edges, once some crucial edges in the optimal solution are missed by the collected candidate edges, it is hard for LKH to reach the global optimum. Simply enlarging the candidate sets allows the algorithm to consider more edges, which, however, reduces the efficiency significantly and may also contain some low-quality edges to mislead the search directions. 


In this paper, we propose a novel method based on reinforcement learning to improve the LKH algorithm, helping it select the candidate edges more smartly and flexibly, providing more edges the opportunity to serve as candidate edges. Specifically, we first expand the candidate sets of each city to a larger size and then select a subset of edges from the candidate sets to serve as candidate edges during each iteration. We associate each city with a multi-armed bandit (MAB), where each arm corresponds to an edge in its enlarged candidate set and is associated with an evaluation value, denoted as $M$-value. The $M$-value of an arm indicates the benefit of selecting it as a candidate edge. The bandit models can learn from the searching process and update the $M$-values. In each iteration of the algorithm, the bandit models recommend appropriate candidate edges for the local search.


Moreover, we employ three different strategies for selecting the candidate edges, i.e., the arms of the bandit. The first strategy applies the $\epsilon$-greedy method to trade-off exploration and exploitation, while the other two make the selection greedily based on the $M$-values and $\alpha$-values, respectively. The three strategies are used cooperatively during the search to improve the robustness.

We apply our proposed method to LKH, denoting the resulting algorithm as 
Bandit-based LKH (\name). Our method expands the searching space, providing higher flexibility for the search, assisting the LKH algorithm in escaping local optima and finding better solutions. We compare \name with the LKH baseline, as well as VSR-LKH~\cite{vsr_lkh} and NeuroLKH~\cite{neuro_lkh}, two representative learning-based algorithms. Extensive experiments in various benchmarks show the superiority of \name over LKH, VSR-LKH, and NeuroLKH. 
In particular, our method exhibits better performance and robustness than simply enlarging the candidate sets in LKH. 
The results indicate that our MAB method can effectively and smartly suggest high-quality edges and ignore low-quality ones.

We further apply our MAB method to the LKH-3 algorithm~\cite{LKH3}, an extension of LKH that can solve 
many TSP variant problems efficiently. The resulting algorithm is denoted as \name-3. We select two representative variant problem called Multiple Traveling Salesmen Problem (MTSP) and Capacitated Vehicle Routing Problem (CVRP), and compare our \name-3 with LKH-3. The results show that our method can also significantly improve LKH-3 in MTSP and CVRP, indicating its excellent generalization capability. 


The main contributions of this paper are as follows.

\begin{itemize}
\item We identified relatively rigid designs in the LKH algorithm, i.e., the candidate edges for each city are relatively fixed. We construct MAB models to select high-quality candidate edges adaptively during the searching process from enlarged candidate sets. We propose three policies to choose candidates coordinately and a framework to train and utilize the MAB models. Our proposed methods and framework can be applied to other heuristic algorithms for routing problems needing to select candidate edges. 



\item We compare \name with LKH as well as representative learning-based algorithms for TSP on various benchmarks. We also generalize our method to LKH-3, which is an extension version of LKH for various TSP variants. Extensive experiments show the excellent performance and generalization capability of our method in TSP and its variant problems, MTSP and CVRP.

\end{itemize}

\section{Problem Definition}
\label{Sec-Prob}
In this section, we present the definition of the involved problems, 
including the Traveling Salesman Problem (TSP), the Multiple TSP (MTSP), and the Capacitated Vehicle Routing Problem (CVRP).

\subsection{Traveling Salesman Problem}
Given an undirected complete graph G(V,E), V is made up of cities in G numbered {1, 2, \ldots, n} and E contains all pairwise edges such as edge between city i and j presented by (i, j) which has its cost d(i,j). The aim is to find an hamiltonian circuit with a minimum total cost caculated by $\sum_{i = 1}^{n-1} (d(i, i+1)) + d(n, 1)$.

\subsection{Multiple TSP}
In the MTSP, a set of $m$ salesmen (or vehicles) are available to visit a set of $n$ cities and each salesman starts and ends their route at a designated depot, with each city needing to be visited exactly once by one of the salesmen. Let $S = \{1, 2, \ldots, m\}$ be the set of salesmen and $C = \{1, 2, \ldots, n\}$ be the set of cities. $T_1, T_2, \ldots, T_m$ are composed of $m$ subsets $C_1, C_2, \ldots, C_m$ in $C$. The goal is to minimize $\sum_{k=1}^{m} \sum_{(i,j) \in T_k} d(i, j)$.


\subsection{Capacitated Vehicle Routing Problem}
In the CVRP, define m homogeneous fleets serving n cities have a max capacity $c$ which could not be exceeded in each vehicle. Due to capacity, n cities are divided into m exclusive parts represented by $T_{1}, T_{2}, \ldots, T_{m}$ besides common starting and ending cities. Each customer in $T_{i}$ is visited exactly once by one of the vehicles and common cities can be visited many times. The total distance traveled by all vehicles, $\sum_{k=1}^{m} \sum_{(i,j) \in T_k} d(i, j)$, is minimized.

\section{Related Work}
\label{Sec-RW}

In this section, we first review some related studies that solve TSP with learning-based methods and then briefly introduce some key components used in LKH.

\subsection{Learning-based Methods}
Recently, the application of learning-based methods, such as reinforcement learning and deep learning, for combinatorial optimization problems has become a highlight of research. TSP has even become a touchstone and one of the most popular problems for learning-based methods.

Learning-based methods for TSP can be roughly divided into two categories depending on whether deep learning is adopted. Actually, most learning-based methods try to apply deep learning methods for TSP solving. Typical methods include deep reinforcement learning and supervised learning. For instance, some studies design or apply novel deep learning structures for solving TSP, such as the multi-pointer Transformer architecture called Pointerformer~\cite{multi_pointer} and the hierarchical deep reinforcement learning framework~\cite{htsp}. 
Some studies train neural networks to automate the heuristic design, such as the Ant Colony Optimization and $k$-opt~\cite{Costa2020,Sui2021,ma2024learning} heuristics. The NeuroLKH algorithm utilizes a Sparse Graph Network 
to select candidate edges for LKH~\cite{neuro_lkh}, showing excellent performance in instances with less than 6,000 cities. Aigerim propose a hybrid model combining an attention-based encoder and a Long Short-Term Memory (LSTM) decoder addressing the challenge of routing a heterogeneous fleet. \cite{drl_Aigerim}. Bresson trains via deep reinforcement learning, effectively addresses the combinatorial optimization challenges of the TSP\cite{dl_transformer}. GRLOS-M uses Graph Neural Networks to improve Adaptive Large Neighbourhood Search~\cite{GRLOS-M}. There are also some other famous models and algorithms using deep learning for TSP, such as S2V-DQN~\cite{Khalil2017}, POMO~\cite{pomo}, GLOP~\cite{glop}, etc.

The deep learning-based methods provide many new and interesting perspectives for solving TSP. However, they are usually hard to scale to large instances with tens of thousands of cities. The other category of method usually uses traditional reinforcement learning to accumulate the learning information in tables and use them to guide the search.traditional reinforcement learning plays an important part role in TSP. Mazyavkina summarized the TSP problem based on reinforcement learning\cite{tradition_rl}. For example, the Ant-Q~\cite{Ant-Q} and Q-ACS~\cite{Q-ACS} replace the pheromone in the ant colony algorithm with the Q-table in the Q-learning algorithm. The RMGA algorithm~\cite{RMGA} uses reinforcement learning to construct mutation individuals in genetic algorithms. RLHEA combines Q-learning method with hybrid evolutionary algorithm for routing problem~\cite{RLHEA}. Moreover, the VSR-LKH~\cite{vsr_lkh} algorithm combines typical reinforcement learning methods with LKH, altering and reordering the candidate edges, showing excellent performance in TSP. Reinforcement learning also has many opportunities in solving stochastic dynamic vehicle routing problem~\cite{opp_sdvrp}.

Our proposed method uses MAB models to select and adjust candidate edges during the local search. Compared to NeuroLKH and VSR-LKH, which also select and determine the candidate edges before the local search process, our method provides more opportunity for other promising edges to be contained in candidate sets, allowing the algorithm to effectively solve instances with various scales and structures. 

\subsection{Key Components in LKH}
\subsubsection{The $\alpha$-value}
Candidate edges are very important for LK-based local search algorithms since they decide the new edges that can be added to the maintained solution. LKH proposes the $\alpha$-value for evaluating the edges and selecting the candidate edges. The $\alpha$-value is calculated based on a 1-tree structure~\cite{Held1970}, a variant of the spanning tree. Given a graph $G= (V, E)$, for any vertex $v \in V$, we can generate a 1-tree by first constructing a spanning tree on $V\backslash \{v\}$ and then combining it with two edges from $E$ incident to $v$. The minimum 1-tree is the 1-tree with the minimum length, i.e., the total length of its edges. We denote $L(T)$ as the length of the minimum 1-tree, which is obviously a lower bound of the length of the shortest TSP tour. Moreover, we denote $L(T(i,j))$ as the length of the minimum 1-tree containing edge $(i,j)$. 
The $\alpha$-value of edge $(i,j)$ is calculated as follows.
\begin{equation}
\alpha(i,j) = L(T(i,j)) - L(T).
\end{equation}

To further enhance the performance of $\alpha$-values, LKH applies the method of adding penalties~\cite{Held1971} to vertices to obtain a tighter lower bound.

\begin{figure}[t]
\centering
\subfigure[Sequential 3-opt]{
\includegraphics[width=0.45\columnwidth]{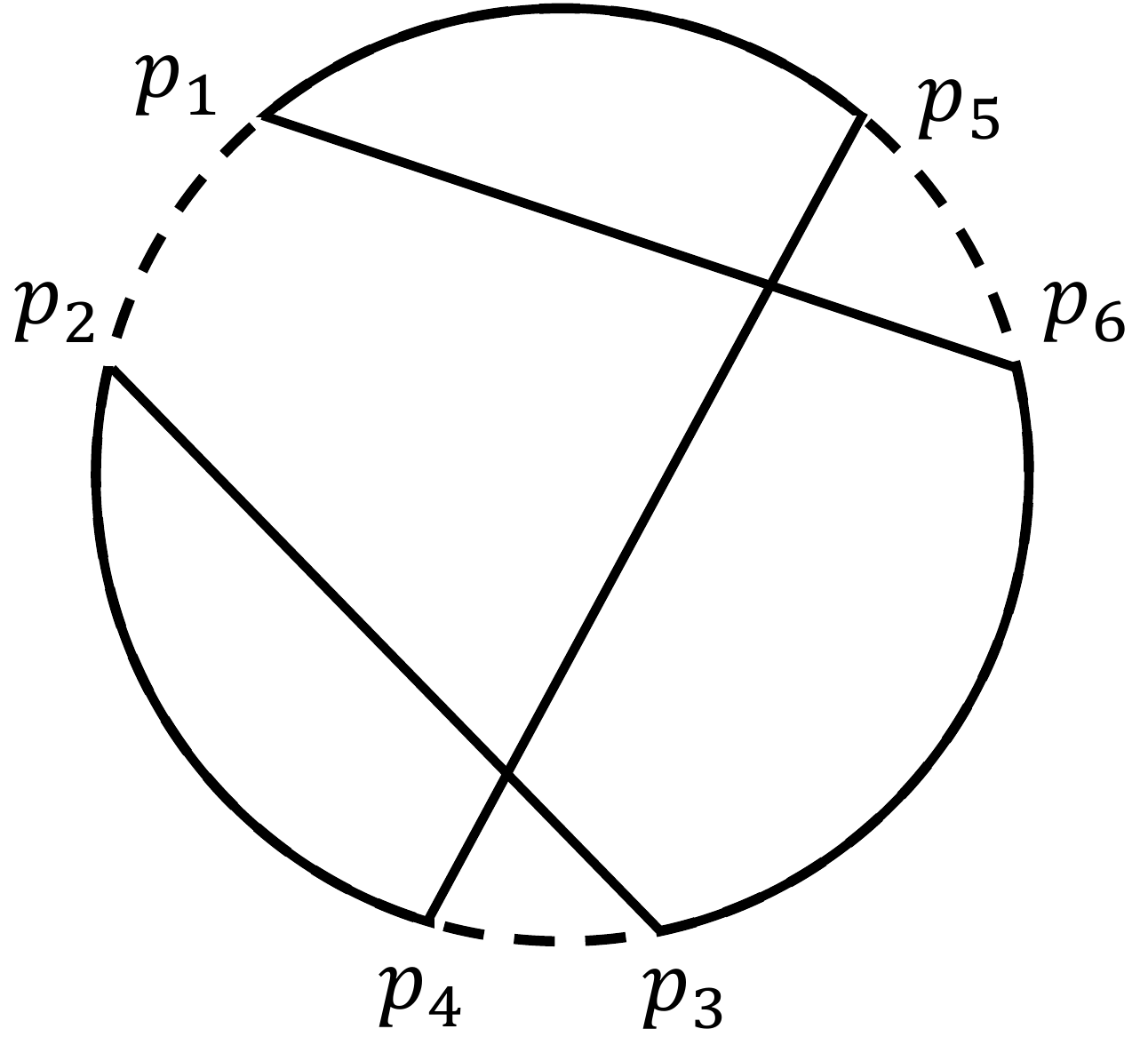}
\label{3-opt}}
\subfigure[Non-sequential 4-opt]{
\includegraphics[width=0.47\columnwidth]{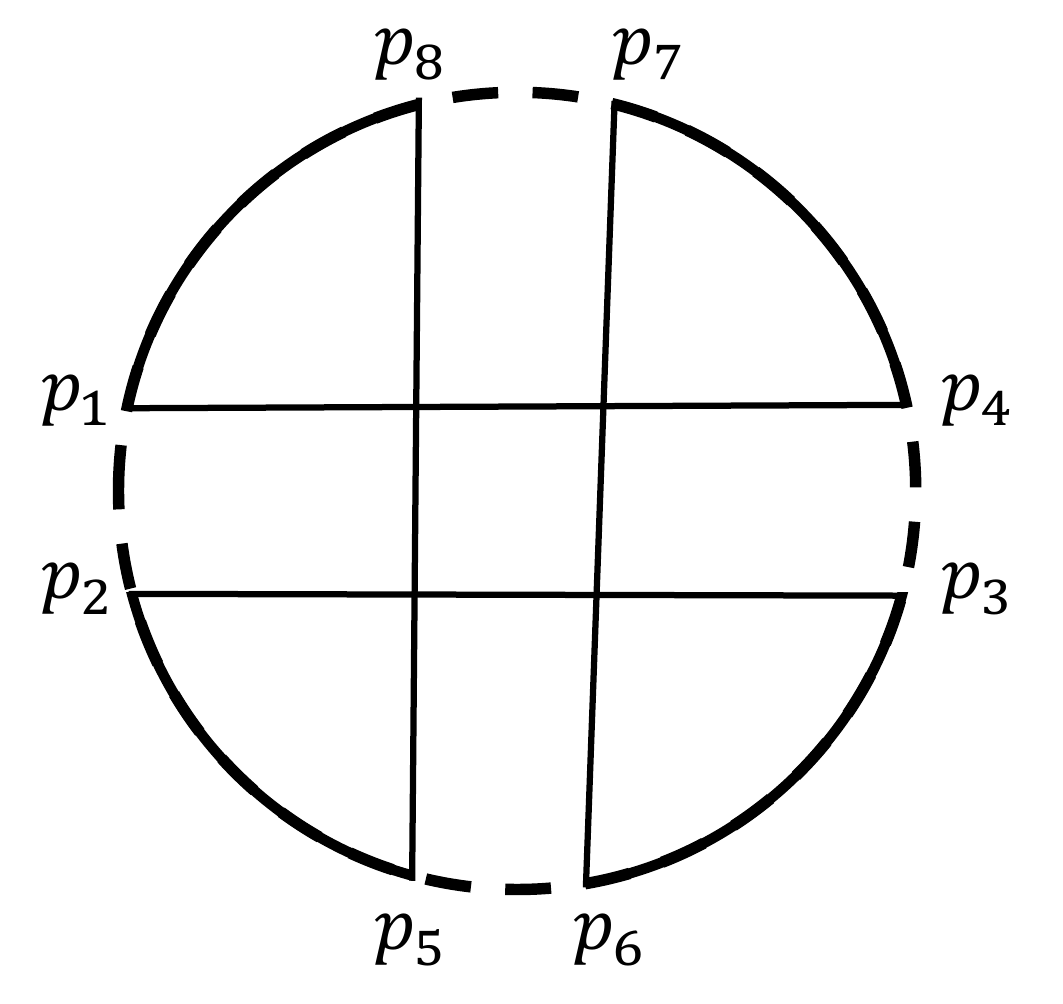}
\label{4-opt}}
\caption{Examples of sequential and non-sequential $k$-opt moves.\vspace{-2.0em}}
\label{34-opt}
\end{figure}


\subsubsection{Search Operators}
\label{sec-searchoperator}
The core search operator in LKH is the famous $k$-opt method, which replaces $k$ edges in the current solution with $k$ new edges based on the eject chain method. That is, remove $k$ edges in the tour and attempt to rearrange the order of connections among these points to find a better solution. The $k$-opt operator in LKH contains two categories, sequential and non-sequential moves, as shown in Figure \ref{34-opt}. The dashed line is the edge that is about to be disconnected. The sequential move starts from a starting point, e.g., $p_1$, alternatively selects the edges to be removed, e.g., $(p_1,p_2)$, $(p_3,p_4)$ and $(p_5,p_6)$, and edges to be added, e.g., $(p_2,p_3)$ and $(p_4,p_5)$, and guarantees that after selecting each edge to be removed, connecting its endpoint, e.g., $p_4$ and $p_6$, back to the starting point leads to a feasible TSP tour. Therefore, the sequential move can be stopped once an improvement is found, and the non-sequential move cannot. The non-sequential move combines two distinct infeasible $k$-opt moves to form a feasible tour, as shown in Figure \ref{4-opt}. 
It is a supplement of the sequential move, exploring additional search space that sequential moves cannot reach.

In summary, the search operators in LKH are encapsulated in the LinKernighan() function, which is an important part of executing LK algorithm~\cite{lk}, following the guidance of the candidate edges to find moves that can improve the current solution. LinKernighan() finally returns a local optimal solution that cannot be improved by any search operators with the candidate edges. For the detailed implementations of the LinKernighan() function, we referred to~\cite{lkh}.





\section{Method}
\label{Sec-Method}
This section presents our proposed \name algorithm. We first introduce our method of using multi-armed bandits (MAB) to select and recommend candidate edges for the local search algorithm, then introduce the main framework of \name, and finally introduce the method for calculating the rewards and updating the bandit models. 

\subsection{Candidate Edges Selection}
As described in the introduction, our proposed MAB models select high-quality candidate edges from the enlarged candidate sets. 
Note that LKH restricts the maximum edges contained in each candidate set using a constant parameter $C_{max}$, whose default value is 5. Such a default value fits well with the algorithm, as a smaller value significantly reduces the search ability, and a larger value may contain low-quality edges and reduce the search efficiency. However, the restriction of $C_{max} = 5$ may also miss some promising edges. 
Therefore, we propose to enlarge each candidate set to a larger size (setting $C_{max} = 7$ by default), allowing more edges to serve as candidate edges, and further use an MAB to ignore low-quality edges and select appropriate edges in each candidate set.

In the experiments, we perform an empirical analysis to evaluate the benefit of the expansion of candidate sets and a comparison showing that using our MAB to select candidate edges from the enlarged candidate sets is more robust and effective than regarding all edges in the enlarged candidate sets as candidate edges directly, indicating the excellent performance of our method.

In the following, we first introduce the proposed MAB model and then the method of pulling the arms in the bandit models.


\subsubsection{The MAB Model}
Given the enlarged candidate set of each city, directly regarding all its edges as candidate edges will significantly reduce the algorithm efficiency and may contain some low-quality edges. Thus, we need to select some appropriate ones from each enlarged candidate set. Such a task background shares similarities with the MAB model, which also needs to perform selections from multiple candidates and lacks prior knowledge for selecting the best ones in each step. There are also some differences between them, i.e., the candidate set needs to produce multiple candidate edges and the MAB model usually only pulls one arm per step. Therefore, we establish the task as a variant MAB model~\cite{MAB-arms}, which pulls multiple arms each time. Specifically, we associate each city with an MAB, where each arm corresponds to an edge in its enlarged candidate set, and pulling each arm corresponds to select the corresponding edge as a candidate edge.

For each arm in each MAB, we assign it an evaluation value, denoted as $M$-value. The $M$-values are initialized to be 0 in the beginning. The larger the $M$-value of an arm, the more the benefit of selecting it as a candidate edge. Our MAB models can learn from the search process and adjust the $M$-values, using them to recommend appropriate candidate edges for the local search algorithm.

\subsubsection{Methods for Pulling the Arms}
\label{pull_arms}
In our method, each MAB needs to pull $N_{arm}$ (5 by default equals to $C_{max}$ of LKH) arms per calling and recommends $N_{arm}$ corresponding candidate edges to participate in the local search process. We design three methods for selecting the arms to be pulled, denoted as $\epsilon$-greedy, $M$-greedy, and $\alpha$-greedy, respectively. Their formal descriptions are as follows.

\begin{itemize}
\item \textbf{$\epsilon$-greedy} randomly explores the edges in the enlarged candidate set with a probability of $\epsilon$ ($0 < \epsilon < 1$) and makes a greedy selection preferring edges with larger $M$-values with a probability of $1-\epsilon$. 
\item \textbf{$M$-greedy} selects $N_{arm}$ edges with the largest $M$-values greedily among each enlarged candidate set.
\item \textbf{$\alpha$-greedy} selects $N_{arm}$ edges with the largest $\alpha$-values greedily among each enlarged candidate set.
\end{itemize}

The above three methods are used alternatively during the search process. Actually, the $M$-values and $\alpha$-values share similarities and differences. Both of them serve as metrics of the edges. The $\alpha$-values are predetermined before the local search process, evaluating the quality of the edges with a global perspective. In contrast, the $M$-values are continuously learned and updated during the local search process. We believe that these two metrics are complementary in evaluating and selecting the candidate edges, and the utilization of the $M$-greedy and $\alpha$-greedy methods combine their complementarity. Moreover, the $\epsilon$-greedy allows the MAB reinforcement learning models to trade-off exploration and exploitation. Experimental results also demonstrate that the combination of the three methods can improve the robustness of the local search algorithm.


Given the $M$-values and $\alpha$-values of the edges, the parameter $\epsilon$, an integer $bandtype$ for determining the selection of the three methods, and the number of pulled arms per calling of MAB $N_{arm}$, Algorithm~\ref{alg_call} shows a CallBandit() function that calls the MAB models to select the candidate edges for the local search algorithm. The set of all selected candidate edges is denoted as $E_{cand}$.

\begin{algorithm}[!t]
\fontsize{10.1pt}{15}
\caption{CallBandit($\epsilon,bandtype,N_{arm}$)}
\label{alg_call}
\LinesNumbered 
\KwIn{$\epsilon$-greedy parameter: $\epsilon$, integer for determining the method: $bandtype$, number of pulled arms: $N_{arm}$}
\KwOut{Candidate edges $E_{cand}$}
\Switch{bandtype}{
\textbf{case} 0 \textbf{do} select $E_{cand}$ by the $\epsilon$-greedy method\;
\textbf{case} 1 \textbf{do} select $E_{cand}$ by the $M$-greedy method\;
\textbf{case} 2 \textbf{do} select $E_{cand}$ by the $\alpha$-greedy method\;
}
\textbf{return} $E_{cand}$\;
\end{algorithm}

\subsection{Main Process of \name}
After establishing the MAB model and the method for selecting candidate edges, we summarize the main framework of \name in this subsection, which is depicted in Algorithm~\ref{alg_main}. In the initialization stage (lines 1-5), the algorithm initializes the enlarged candidate sets, the $M$-values, and some information involved in the search process. During the local search process (lines 6-19), the algorithm first generates an initial solution $R$ by function ChooseInitialTour() in LKH (line 7) and then calls the MAB models by function CallBandit() (Algorithm~\ref{alg_call}) to select the candidate edges $E_{cand}$ (line 8). The selected $E_{cand}$ will be used to guide the LinKernighan() function, which performs the search operators introduced in Section \ref{sec-searchoperator} to improve $R$ to a local optimum (line 9). The $M$-values of the selected candidate edges will be updated according to the quality of the obtained local optimal solution (line 10). Moreover, if the best solution $R^*$ has not been improved for $T_{type}$ trials, the $bandtype$ will be changed (line 18) to try to use different arm selection methods to help the algorithm escape from local optima. 

For the time complexity, the additional operations of bandLKH over LKH are the CallBandit and UpdateM functions. CallBandit sorts the enlarged candidate set with size $C_{max}$ (7 by default) of each city, costing a time complexity of $O(nC_{max}log(C_{max})$, where $n$ is the number of cities. UpdateM traverses the selected candidate edges once, costing a time complexity of $O(n)$. For the space complexity, bandLKH associates an $M$-value with each candidate city, costing a space complexity of $O(C_{max}n)$. Therefore, both the time and space complexities are of linear order and acceptable for large-scale problems.

In summary, the proposed \name algorithm uses the MAB models to help select candidate edges for the local search process. The MAB models select the candidate edges referring to the evaluation values of the arms, i.e., $M$-values, and can learn from the searching process to update the $M$-values and recommend appropriate and high-quality candidate edges. The selected candidate edges can be used for various search operators in LKH encapsulated in the LinKernighan() function, including sequential and non-sequential moves.

\begin{algorithm}[!t]
\fontsize{10.1pt}{15}
\caption{\name}
\label{alg_main}
\LinesNumbered 
\KwIn{A TSP instance: $I$, enlarged size of candidate sets: $C_{max}$, maximum number of trials: $MaxTrials$, the cut-off time: $MaxTime$, number of pulled arms: $N_{arm}$, maximum no-improvement trials for change $bandtype$: $T_{type}$, $\epsilon$-greedy parameter: $\epsilon$, incremental reward parameter: $\lambda$}
\KwOut{The best solution found for $I$: $R^*$}
Initialize each enlarged candidate set containing $C_{max}$ edges based on $\alpha$-values\;
Initialize length of the best solution $L(R^*) \leftarrow +\infty$\;
Initialize the number of no-improvement trials $t \leftarrow 0$\; Initialize $bandtype \leftarrow 0$\;
Initialize the $M$-values to 0\;
\For{$i \leftarrow 1 : MaxTrials$}{
    $R \leftarrow$ ChooseInitialTour()\;
    $E_{cand} \leftarrow$ CallBandit($\epsilon,bandtype,N_{arm}$)\;
    $R \leftarrow$ LinKernighan($I,R,E_{cand}$)\;
    UpdateM($E_{cand},R,R^*,\lambda$)\;
    \eIf{$L(R) < L(R^*)$}{
        $R^* \leftarrow R$\;
        $t \leftarrow 0$\;
    }
    {
        $t \leftarrow t + 1$\;
        \If{$t \geq T_{type}$}{
        $t \leftarrow$ 0\; 
        $bandtype \leftarrow (bandtype + 1)\% 3$\;
        }
    }
    \lIf{running time $>MaxTime$}{\textbf{break}}
}
\textbf{return} $R^*$\;
\end{algorithm}

\subsection{Update the $M$-values with Rewards}
To update the evaluation values, i.e., $M$-values, of the arms in our MAB models, we need to design the reward functions for evaluating the benefit of pulling the arms, i.e., selecting the corresponding candidate edges. Suppose $R^*$ is the best solution found during the search process and $R$ is the solution obtained by the function LinKernighan() based on the selected candidate edges in the current trial. A simple and straightforward reward function is designed as follows.




\begin{equation}
\label{EQ_r}
    r(R,R^*) = L(R^*) - L(R).
\end{equation}

Given the above reward function, the function UpdateM() updates the $M$-value of each selected candidate edge $(i,j) \in E_{cand}$ in an incremental manner by the following equation.

\begin{equation}
\label{EQ_rl}
M(i,j) = (1 - \lambda) \cdot M(i,j) + \lambda \cdot r(R,R^*),
\end{equation}
where $0 < \lambda < 1$ is the incremental reward parameter.



\section{Experiment}
\label{Sec-Exp}
\label{Exp}
In this section, we first make a detailed comparison between \name 
and LKH\footnote{http://webhotel4.ruc.dk/~keld/research/LKH/} (version 2.0.10) to evaluate the performance of our proposed new algorithm. We also compare \name with representative learning-based methods, including the deep learning-based NeuroLKH algorithm~\cite{neuro_lkh} and the traditional reinforcement learning-based VSR-LKH algorithm~\cite{vsr_lkh}. We further present ablation studies to 
evaluate the efficacy of components in \name and provide further insights. 

Finally, we apply our method to LKH-3~\cite{LKH3}, an extension of LKH that can solve many 
TSP variants. 
The resulting solver is called \name-3. We select two representative variants of TSP: the Multiple Traveling Salesmen Problem (MTSP), where the cities are visited by $m$ salesmen, and the goal is to minimize their total traveling distance, and the Capacitated Vehicle Routing Problem (CVRP), which regards the salesmen and cities as vehicles with capacities and customers with demands, and the goal is to minimize the total traveling distance of the vehicles while ensuring that each customer's demand is satisfied and the total demand does not exceed the capacity of any vehicle. We compare \name-3 with LKH-3 in solving MTSP and CVRP and evaluate the generalization capability of our method.


\begin{table*}[!t]
\caption{Detailed comparison results of \name and LKH. The best results appear in bold.\vspace{0.5em}}
\centering
\footnotesize
\resizebox{\linewidth}{!}{

\begin{tabular}{lrrrrrrrrrrrr}  \toprule
\multirow{2}{*}{Instance} & \multirow{2}{*}{BKS} & \multicolumn{5}{c}{LKH}                                                                              & \multicolumn{1}{c}{} & \multicolumn{5}{c}{\name}                                                          \\  \cline{3-7} \cline{9-13} 
                          &                      & Success              & Best (\%)                  & Average (\%)                & Trials  & Time (s) &                      & Success        & Best (\%)                   & Average (\%)                  & Trials  & Time (s) \\   \hline
gr137                     & 69853                & \textbf{10/10}       & \textbf{69853 (0.0000)}    & \textbf{69853 (0.0000)}     & 1.0     & 0.01     &                      & 9/10           & \textbf{69853 (0.0000)}     & 69865.7 (0.0182)              & 18.9    & 0.01     \\
ch150                     & 6528                 & 9/10                 & \textbf{6528 (0.0000)}     & 6528.5 (0.0077)             & 60.1    & 0.05     & \textbf{}            & \textbf{10/10} & \textbf{6528 (0.0000)}      & \textbf{6528 (0.0000)}        & 23.3    & 0.08     \\
kroB150                   & 26130                & 2/10                 & \textbf{26130 (0.0000)}    & 26131.6 (0.0061)            & 128.4   & 0.24     & \textbf{}            & \textbf{10/10} & \textbf{26130 (0.0000)}     & \textbf{26130 (0.0000)}       & 54.1    & 0.21     \\
si175                     & 21407                & 7/10                 & \textbf{21407 (0.0000)}    & 21407.3 (0.0014)            & 105.9   & 6.15     & \textbf{}            & \textbf{9/10}  & \textbf{21407 (0.0000)}     & \textbf{21407.1 (0.0005)}     & 35.3    & 4.95     \\
rat195                    & 2323                 & 9/10                 & \textbf{2323 (0.0000)}     & 2323.5 (0.0215)             & 55.0    & 0.15     & \textbf{}            & \textbf{10/10} & \textbf{2323 (0.0000)}      & \textbf{2323 (0.0000)}        & 14.8    & 0.19     \\
\bottomrule     
\end{tabular}

}
\vspace{0.5em}
\label{table-main}
\end{table*}

\begin{table*}[!t]
\caption{Summarized comparison results of \name and learning-based methods, VSR-LKH and NeuroLKH. The best results appear in bold.\vspace{0.5em}}
\centering
\footnotesize

\begin{tabular}{l|rr|rr|rr}     \toprule
Indicator & \name & VSR-LKH  & \name & NeuroLKH\_R & \name & NeuroLKH\_M \\ \hline
$Win_{best}$       & \textbf{2}           & 0        & \textbf{8}           & 0           & \textbf{5}           & 0           \\
$Win_{avg}$       & \textbf{13}          & 9        & \textbf{22}          & 8           & \textbf{13}          & 9           \\
$Gap_{best}$       & \textbf{0.0000\%}    & 0.0027\% & \textbf{0.0000\%}    & 0.0215\%    & \textbf{0.0000\%}    & 0.0803\%    \\
$Gap_{avg}$       & \textbf{0.0036\%}    & 0.0060\% & \textbf{0.0043\%}    & 0.0366\%    & \textbf{0.0043\%}    & 0.0834\%     \\ \bottomrule   
\end{tabular}
\vspace{0.5em}

\label{table-vsr}
\end{table*}



\begin{table*}[!t]
\caption{
Summarized comparison results of \name and its variants, \name-no$\epsilon$, \name-no$M$, and \name-no$\alpha$. The best results appear in bold.\vspace{0.5em}}
\centering
\footnotesize
\begin{tabular}{l|rr|rr|rr} \toprule
Indicator         & \name     & \name-no$\epsilon$   & \name     & \name-no$M$   & \name     & \name-no$\alpha$ \\ \hline
$Win_{best}$ & 0        & 0        & 0       & 0        & \textbf{2}        & 0          \\ 
$Win_{avg}$  & \textbf{12}       & 7        & 6       & \textbf{12}        & \textbf{28}       & 4          \\ 
$Gap_{best}$ & \textbf{0.0000\%} & \textbf{0.0000\%} & \textbf{0.0000\%} & \textbf{0.0000\%} & \textbf{0.0000\%} & 0.0006\%   \\ 
$Gap_{avg}$  & \textbf{0.0036\%} & 0.0044\% & \textbf{0.0036\%} & 0.0057\% & \textbf{0.0036\%} & 0.0076\%   \\ \bottomrule
\end{tabular}
\vspace{0.5em}

\label{table-ablation1}
\end{table*}




\begin{table}[!t]
\caption{Summarized comparison results between \name with LKH-$N_{arm}$ and LKH-$C_{max}$. The best results appear in bold.\vspace{0.5em}}
\centering
\footnotesize
\begin{tabular}{l|rr} \toprule
Indicator        & \name     & LKH-$C_{max}$ \\ \hline
$Win_{best}$   & \textbf{1}        & 0 \\       
$Win_{avg}$       & \textbf{19}       & 12      \\
$Gap_{best}$ & \textbf{0.0000\%} & 0.0001\% \\
$Gap_{avg}$  & \textbf{0.0036\%} & 0.0057\%  \\ \bottomrule
\end{tabular}
\vspace{0.5em}

\label{table-ablation2}
\end{table}

\begin{table*}[!t]
\caption{Detailed comparison of LKH3 and \name-3 on CVRP. The best results appear in bold.\vspace{0.5em}}
\centering
\footnotesize

\begin{tabular}{lrrrrrrrrr} \toprule
            & \multicolumn{4}{c}{LKH-3}                                                                                         &                      & \multicolumn{4}{c}{\name-3}                                                               \\   \cline{2-5} \cline{7-10}
Instance   & \multicolumn{1}{r}{Best} & \multicolumn{1}{r}{Average} & \multicolumn{1}{r}{Trials} & \multicolumn{1}{r}{Time(s)} & \multicolumn{1}{r}{} & Best            & \multicolumn{1}{r}{Average} & \multicolumn{1}{r}{Trials} & \multicolumn{1}{r}{Time(s)} \\    \hline
X-n101-k25  & 27653                    & 27717.4                     & 10000.0                    & 56.48                       &                      & \textbf{27591}  & \textbf{27625}              & 7393.0            & 19.872                      \\
X-n106-k14  & 26389                    & 26482.2                     & 10000.0                    & 50.67                       &                      & \textbf{26386}  & \textbf{26388.2}            & 10000.0           & 61.72                       \\
X-n110-k13  & \textbf{14971}           & \textbf{14971}              & 1271.0                     & 0.37                        &                      & \textbf{14971}  & \textbf{14971}              & 2048.0            & 0.698                       \\
X-n115-k10  & \textbf{12747}           & 12754                       & 5159.0                     & 1.76                        &                      & \textbf{12747}  & \textbf{12747}              & 314.0             & 0.069                       \\
X-n120-k6   & \textbf{13332}           & \textbf{13351.5}            & 6011.0                     & 3.03                        &                      & 13340           & 13374.4                     & 10000.0                    & 12.692                      \\
\bottomrule           
\end{tabular}

\vspace{0.5em}
\label{appendix_cvrp1}
\end{table*}

\begin{table*}[!t]
\caption{Detailed comparison of LKH3 and \name-3 on CVRP (continued). The best results appear in bold.\vspace{0.5em}}
\centering
\footnotesize

\begin{tabular}{lrrrrrrrrr} \toprule
            & \multicolumn{4}{c}{LKH-3}                                                                                         &                      & \multicolumn{4}{c}{\name-3}                                                               \\   \cline{2-5} \cline{7-10}
Instance   & \multicolumn{1}{r}{Best} & \multicolumn{1}{r}{Average} & \multicolumn{1}{r}{Trials} & \multicolumn{1}{r}{Time(s)} & \multicolumn{1}{r}{} & Best            & \multicolumn{1}{r}{Average} & \multicolumn{1}{r}{Trials} & \multicolumn{1}{r}{Time(s)} \\    \hline
X-n317-k53  & \textbf{78389}           & \textbf{78608.6}            & 10000.0                    & 78.26                       &                      & 78458           & 78679.6                     & 10000.0                    & 106.873                     \\
X-n322-k28  & \textbf{30228}           & 30369.4                     & 10000.0                    & 72.03                       &                      & 30233           & \textbf{30263.6}            & 10000.0                    & 83.37                       \\
X-n327-k20  & \textbf{27887}           & \textbf{28015.8}            & 10000.0                    & 58.75                       &                      & 28027           & 28078.8                     & 10000.0                    & 73.938                      \\
X-n331-k15  & 31488                    & \textbf{31565}              & 10000.0                    & 30.20                       &                      & \textbf{31393}  & 31608.8                     & 10000.0                    & 40.726                      \\
X-n336-k84  & \textbf{140006}          & 140344.8                    & 10000.0                    & 231.94                      &                      & 140023          & \textbf{140283.4}           & 10000.0                    & 282.161                     \\
\bottomrule
\end{tabular}

\vspace{0.5em}
\label{appendix_cvrp2}
\end{table*}

\begin{table*}[!t]
\caption{Detailed comparison of LKH3 and \name-3 on MTSP. The best results appear in bold.\vspace{0.5em}}
\centering
\footnotesize
\resizebox{\linewidth}{!}{

\begin{tabular}{lrrrrrrrrrlrrrrrrrr}    \toprule
\multirow{2}{*}{Instance} & \multicolumn{4}{c}{LKH3}                                & \multicolumn{4}{c}{\name-3}                           & \multicolumn{1}{r}{} & \multirow{2}{*}{Instance} & \multicolumn{4}{c}{LKH3}                                   & \multicolumn{4}{c}{\name-3}                                  \\ \cline{2-9} \cline{12-19}
                           & Best            & Average           & Trials & Time (s) & Best            & Average           & Trials & Time (s) & \multicolumn{1}{r}{} &                            & Best              & Average           & Trials  & Time (s) & Best               & Average              & Trials  & Time (s) \\ \hline
eil101                     & \textbf{646}    & \textbf{646}      & 128.0  & 0.00     & \textbf{646}    & \textbf{646}      & 114.0  & 0.00     &                      & u574                       & \textbf{37012}    & \textbf{37113}    & 1000.0  & 0.09     & 37030              & 37168                & 1000.0  & 0.11     \\
lin105                     & \textbf{14741}  & \textbf{14741}    & 1.0    & 0.00     & \textbf{14741}  & \textbf{14741}    & 1.0    & 0.00     &                      & rat575                     & \textbf{6821}     & \textbf{6850.6}   & 1000.0  & 0.09     & 6832               & \textbf{6850.6}      & 1000.0  & 0.11     \\
pr107                      & \textbf{44869}  & \textbf{44869}    & 1.0    & 0.00     & \textbf{44869}  & \textbf{44869}    & 1.0    & 0.00     &                      & p654                       & \textbf{36890}    & 37413.6           & 975.2   & 0.13     & \textbf{36890}     & \textbf{36897.67}    & 800.3   & 0.10     \\
\bottomrule
\end{tabular}

}
\vspace{0.5em}
\label{appendix_mtsp1}
\end{table*}

\subsection{Experimental Setup}
\label{setup}
\name was implemented in C Programming Language. The experiments were run on a server using an AMD EPYC 7H12 CPU, running Ubuntu 18.04 Linux operation system. We tested the algorithms in all the 82 symmetry TSP instances from the TSPLIB\footnote{http://comopt.ifi.uni-heidelberg.de/software/TSPLIB95/} benchmark with the number of cities ranging from 101 to 85,900, and all the 22 instances with the number of cities ranging from 10,000 to 30,000 from the National TSPs\footnote{https://www.math.uwaterloo.ca/tsp/world/countries.html} and VLSI TSPs\footnote{https://www.math.uwaterloo.ca/tsp/vlsi/index.html} benchmarks. We also use the 82 instances from TSPLIB with $m = 3$ as the MTSP benchmarks. A TSP instance can be transformed into an MTSP instance by regarding the first city as the depot. Note that the number in the name of a TSP instance indicates the number of cities it contains. For CVRP, we tested LKH-3 and \name-3 in the dataset proposed by Uchoa et al.~\cite{UchoaPPPVS17}\footnote{http://www.vrp-rep.org/datasets/item/2016-0019.html} containing 100 instances with the number of customers ranging from 100 to 1,001.

We set the cut-off time $MaxTime$ and the maximum number of iterations $MaxTrials$ to be the same for the algorithms. $MaxTrials$ is set to the number of cities (default settings in LKH and LKH-3) for instances with less than 10,000 cities, 10,000 for instances with more than 10,000 cities, and 3,000 for instances with more than 30,000 cities. 
Each TSP instance with less than 10,000 cities was run 10 times by each algorithm with different random seeds. The rest of the instances were run 5 times by each algorithm. $MaxTime$ is set to one day for all the instances. 
In each run, the algorithm will terminate and start the next run when it finds the optimal solution.

Parameters related to the MAB in \name and \name-3 include the enlarged size of candidate sets: $C_{max}$, the number of pulled arms in each bandit: $N_{arm}$, the maximum no-improvement trials for change $bandtype$: $T_{type}$, the $\epsilon$-greedy parameter: $\epsilon$, and the incremental reward parameter: $\lambda$. We adopted an automatic configurator called SMAC3~\cite{SMAC} to tune them based on some sampled instances, and their default settings are as follows: $C_{max} = 7$, $N_{arm} = 5$, $T_{type} = MaxTrials / 20$ by referring to VSR-LKH, $\epsilon = 0.15$, and $\lambda = 0.16$. Other parameters are the same as those in LKH and LKH-3. The tuning domains of $C_{max}$, $\epsilon$, and $\lambda$ are $[6,8]$, $[0.05, 0.06, ..., 0.30]$, and $[0.05, 0.06, ..., 0.30]$, respectively.

\subsection{Comparison of \name and LKH}
\label{main_exep}
This subsection compares \name and LKH. Actually, LKH selects 5 fixed candidate edges for each city, and \name selects $N_{arm} = 5$ more appropriate candidate edges in a larger candidate set (with size equals $C_{max} = 7$) for each city using the proposed MAB model.

The detailed comparison results between \name and LKH are shown in Table~\ref{table-main}. Column $BKS$ indicates the best-known solutions of the instances, and we only present all the 62 instances in which at least one of LKH and \name cannot obtain the $BKS$ in each run. Actually, the $BKS$ of tested instances from TSPLIB and National TSPs have been proven to be optimal. Column \textit{Success} indicates the success rate to find the $BKS$, columns \textit{Best} and \textit{Average} indicate the best and average solutions of the algorithms, respectively, with their gaps to the $BKS$ placed in the brackets, and columns \textit{Trials} and \textit{Time} indicate the average trials and running time of the algorithms. We further present the average gap of the best and average solutions to the $BKS$ at the bottom of Table~\ref{table-main}.

The results show that \name obtains better (resp. worse) results in terms of the success rate in 33 (resp. 1) instances and better (resp. worse) results in terms of the average solution in 52 (resp. 10) instances. The average gap of the best (resp. average) solutions obtained by \name to the $BKS$ is about 65\% (resp. 57) smaller than that of LKH. We can also observe that \name significantly outperforms LKH in solving instances with various scales. The results indicate that \name has significantly better performance and robustness than LKH. 



\subsection{Comparison with Learning-based Methods}
To further show the performance and advantage of our learning method, we compare \name with the state-of-the-art learning-based methods for TSP, including the NeuroLKH algorithm~\cite{neuro_lkh}, which combines deep learning and LKH, and the VSR-LKH algorithm~\cite{vsr_lkh}, which combines traditional reinforcement learning with LKH. We compare \name with two versions of NeuroLKH, NeuroLKH\_R and NeuroLKH\_M, which were trained in instances with uniformly distributed nodes and a mixture of instances with uniformly distributed nodes, clustered nodes, half uniform and half clustered nodes, respectively. We compare \name with VSR-LKH in 75 TSPLIB instances whose number of cities ranges from 101 to 10,000 and compare \name with NeuroLKH in 60 TSPLIB instances reported in its paper. The summarized results are all shown in Table~\ref{table-vsr}. 

As shown in Table~\ref{table-vsr}, 
\name obtains better results than VSR-LKH in 2 (resp. 13) instances in terms of the best (resp. average) solutions and worse results than VSR-LKH in 0 (resp. 9) instances in terms of the best (resp. average) solutions. 
Moreover, \name obtains better results than NeuroLKH\_R in 8 (resp. 21) instances in terms of the best (resp. average) solutions and worse results than NeuroLKH\_R in 1 (resp. 8) instances in terms of the best (resp. average) solutions, and \name obtains better results than NeuroLKH\_M in 5 (resp. 13) instances in terms of the best (resp. average) solutions and worse results than NeuroLKH\_M in 1 (resp. 9) instances in terms of the best (resp. average) solutions. 

The results in Table~\ref{table-vsr} indicate that \name has a superiority over the state-of-the-art learning-based methods, VSR-LKH and NeuroLKH. Note that NeuroLKH uses deep learning methods to help LKH select candidate edges before local search, and VSR-LKH uses traditional reinforcement learning to adjust the order of the candidate edges during the search. Therefore, neither of them changes the candidate edges of each city. The proposed \name algorithm can dynamically suggest candidate edges from enlarged candidate sets, providing more flexibility for the algorithm and leading the algorithm to escape from local optima effectively. Thus, our learning method shows better performance than existing ones for TSP.

\subsection{Ablation Study}
\label{ablation_exp}
We perform two groups of ablation studies to evaluate the effect of components in our method. The first group focuses on the arm selection strategies by comparing \name with its three variants, \name-no$\epsilon$, \name-no$M$, and \name-no$\alpha$, which remove the $\epsilon$-greedy, $M$-greedy, and $\alpha$-greedy strategies from \name, respectively.
The second group focuses on the MAB model by comparing \name with a variant algorithm of LKH, LKH-$C_{max}$, which sets the maximum capacity of each candidate to $C_{max}=7$ and regards all the edges of each candidate set as candidate edges. 
The ablation studies are performed based on all the 75 TSPLIB instances whose number of cities ranges from 101 to 10,000.


\subsubsection{Ablation Study on the Arm Selection Strategy}
Due to the limited space, we present summarized comparison results of \name and its variants, \name-no$\epsilon$, \name-no$M$, and \name-no$\alpha$, in Table~\ref{table-ablation1}, where rows $Win_{best}$ and $Win_{avg}$ indicate the number of instances in which the algorithm obtains better results of the best and average solution in multiple runs than its competitor, respectively, and rows $Gap_{best}$ and $Gap_{avg}$ indicate the average values of the gap of the best and average solutions to the optimal solutions upon all the 75 instances, respectively. The detailed results are referred to in the Appendix. The results show that \name generally outperforms the three variants, indicating that our selected three arm selection strategies are effective and complementary, and \name combines their complementarity to improve the robustness.



\subsubsection{Ablation Study on the MAB Model}
The summarized comparison results between \name with LKH-$C_{max}$ are shown in Table~\ref{table-ablation2}. The results show that 
\name exhibits better performance and robustness than LKH-$C_{max}$ in terms of the best and average solutions, indicating that directly enlarging the candidate edges is also not a good choice since there might be some low-quality edges contained, and our MAB model can learn from the search process and smartly select candidate edges and ignore low-quality edges in the enlarged candidate sets, making our \name significantly outperform the LKH algorithm and also show higher robustness than the LKH-$C_{max}$ algorithm without a smart filter.





\subsection{Generalization Evaluation on MTSP and CVRP}
\label{MTSP_CVRP}
Finally, we compare \name-3 with LKH-3 in CVRP and MTSP and full results are presented in Table~\ref{appendix_cvrp1}, Table~\ref{appendix_cvrp2} and Table~\ref{appendix_mtsp1} respectively. As shown in full results corresponding to CVRP, \name-3 shows better performance than LKH-3 in most instances containing X-n101-k25, X-n289-k60, X-n480-k70 and X-n936-k151 \ldots. In MTSP, \name-3 also performs better than LKH-3 in medium and hard instances, such as ali535, rl1304, d2103, rl11849 and pla85900, which shows that \name-3 is suitable for different scale of the problem. As time goes by, our proposed method could learn more beneficial information from passing trials so that guide algorithm to explore more significant search space.


Generally speaking, the results show that among all the 82 tested MTSP instances, \name-3 obtains better results than LKH-3 in 33 (resp. 47) instances in terms of the best (resp. average) solutions and worse results than LKH-3 in 21 (resp. 19) instances in terms of the best (resp. average) solutions. 
Among all the 100 tested CVRP instances, \name-3 obtains better results than LKH-3 in 56 (resp. 61) instances in terms of the best (resp. average) solutions and worse results than LKH-3 in 35 (resp. 35) instances in terms of the best (resp. average) solutions. The results 
also show that \name-3 exhibits excellent performance in MTSP and CVRP instances across various scales. \name-3 exhibits a general improvement over LKH-3 in MTSP and CVRP, and the results indicate the excellent generalization capability of our MAB method. 

\section{Conclusion}
\label{Sec-Con}

Selecting appropriate candidate edges is crucial for solving various routing problems, which decides the new edges that can be used to adjust solutions and explore the solution space. Larger candidate sets indicate wider search space and lower efficiency at the same time. Balancing the search accuracy and efficiency is the magic of heuristic methods. The LKH algorithm achieves this by tuning the candidate set to a reasonable size. In this work, we propose an alternative may, i.e., modeling the selection of candidate edges 
as pulling arms in multi-armed bandits, 
providing opportunities for more potential edges to be considered as candidate edges and learning from the searching process to select appropriate candidate edges. 
Extensive comparisons, ablation studies, and evaluations on generalization capability demonstrate the effectiveness of our proposed algorithm.


Though multi-armed bandit is not a new technology, identifying the 
limitations of LKH and combining the bandit model with the problem context to improve the local search algorithm is indeed of significant research value. 
In the future, we will continue to refine \name, such as adaptively changing the size of the candidate sets to enhance the performance and robustness for problems in various scales. 
Also, our method has the potential to be extended to other routing problems, such as various variants of TSP and vehicle routing problems, that also need to select high-quality candidate edges. 

\section*{Acknowledgments}
This work is supported by National Natural Science Foundation of China (U22B2017) and Microsoft Research Asia (100338928).


\bibliographystyle{unsrt}

\bibliography{main}

\end{document}